\documentclass[10pt,twocolumn,letterpaper]{article}

\usepackage{cvpr}              

\usepackage{graphicx}
\usepackage{color, xcolor}
\usepackage{amsmath}
\usepackage{amssymb}
\usepackage{multirow}
\usepackage{booktabs}
\usepackage{bbm}
\usepackage{bm}
\usepackage[ruled,vlined]{algorithm2e} 

\usepackage{pifont}
\newcommand{\cmark}{\ding{51}}%

\usepackage[pagebackref,breaklinks,colorlinks]{hyperref}

\colorlet{dark-blue}{blue!70!black}
\hypersetup{
    colorlinks=true,%
    citecolor=dark-blue,%
    filecolor=dark-blue,%
    linkcolor=red,%
    urlcolor=magenta
}

\usepackage[capitalize]{cleveref}
\crefname{section}{Sec.}{Secs.}
\Crefname{section}{Section}{Sections}
\Crefname{table}{Table}{Tables}
\crefname{table}{Tab.}{Tabs.}

\usepackage[normalem]{ulem}
\usepackage{comment}

\usepackage{caption}
\def\cvprPaperID{6008} 
\def\confName{CVPR}
\def\confYear{2023}

\begin{document}

\title{Dynamic Conceptional Contrastive Learning for \\ Generalized Category Discovery}

\author{Nan Pu~~~~Zhun Zhong$^*$~~~~Nicu Sebe\\
The Department of Information Engineering and Computer Science\\
University of Trento, Trento, Italy\\
{\tt\small \{nan.pu, zhun.zhong, niculae.sebe\}@unitn.it}
}
\maketitle

\begin{abstract}

    Generalized category discovery (GCD) is a recently proposed open-world problem, which aims to automatically cluster partially labeled data. The main challenge is that the unlabeled data contain instances that are not only from known categories of the labeled data but also from novel categories. This leads traditional novel category discovery (NCD) methods to be incapacitated for GCD, due to their assumption of unlabeled data are only from novel categories. One effective way for GCD is applying self-supervised learning to learn discriminate representation for unlabeled data. However, this manner largely ignores underlying relationships between instances of the same concepts (\textit{e.g., class, super-class, and sub-class}), which results in inferior representation learning. In this paper, we propose a Dynamic Conceptional Contrastive Learning (DCCL) framework, which can effectively improve clustering accuracy by alternately estimating underlying visual conceptions and learning conceptional representation. In addition, we design a dynamic conception generation and update mechanism, which is able to ensure consistent conception learning and thus further facilitate the optimization of DCCL. Extensive experiments show that DCCL achieves new state-of-the-art performances on six generic and fine-grained visual recognition datasets, especially on fine-grained ones. For example, our method significantly surpasses the best competitor by 16.2\% on the new classes for the CUB-200 dataset. Code is available at \url{https://github.com/TPCD/DCCL}
    
\end{abstract}

\section{Introduction}
\label{sec:intro}
\renewcommand{\thefootnote}{\fnsymbol{footnote}} 
\footnotetext[1]{Corresponding Author.}
\begin{figure}[t]
  \begin{center}
  \includegraphics[width=0.48\textwidth]{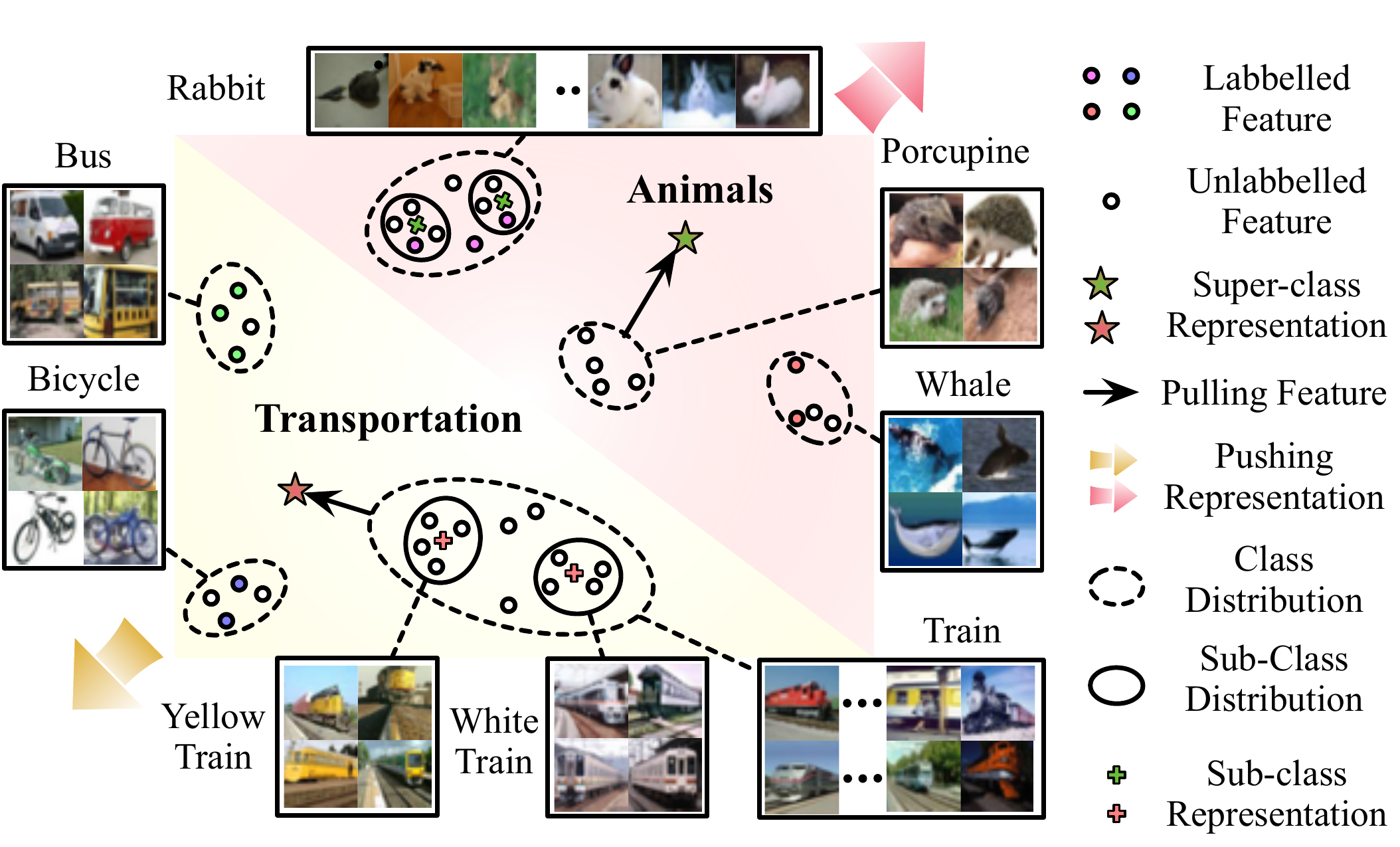}
\caption{Diagram of the proposed Dynamic Conceptional Contrastive Learning (DCCL). Samples from the conceptions should be close to each other. For example, samples from the same classes (bus) at the class level, samples belonging to the transportation (bus and bicycle) at the super-class level, and samples from trains with different colors at the sub-class level. Our DCCL potentially learns the underlying conceptions in unlabeled data and produces more discriminative representations.
\label{fig:conceptual}}
  \end{center}
\vspace{-2em}
\end{figure}

Learning recognition models (\textit{e.g.}, image classification) from labeled data has been widely studied in the field of machine learning and deep learning \cite{he2016deep,krizhevsky2017imagenet,DBLP:journals/corr/SimonyanZ14a}.
In spite of their tremendous success, supervised learning techniques rely heavily on huge annotated data, which is not suitable for open-world applications. 
Thus, the researchers recently have paid much effort on learning with label-imperfection data, such as semi-supervised learning~\cite{oliver2018realistic,sohn2020fixmatch}, self-supervised learning~\cite{zhai2019s4l,he2020momentum}, weakly-supervised learning~\cite{yu2020weakly,zheng2021weakly}, few-shot learning~\cite{snell2017prototypical,wang2020generalizing}, open-set recognition \cite{scheirer2012toward} and learning with noisy labels~\cite{yi2022learning}, etc.

\par

Recently, inspired by the fact that Humans can easily and automatically learn new knowledge with the guidance of previously learned knowledge, novel category discovery (NCD)~\cite{han2019learning, fini2021unified,zhong2021openmix,incd2022,zhao2022ncdss} is introduced to automatically cluster unlabeled data of unseen categories with the help of knowledge from seen categories. However, the implementation of NCD is under a strong assumption that all the unlabeled instances belong to unseen categories, which is not practical in real-world applications. To address this limitation, Vaze \etal \cite{vaze2022generalized} extend NCD to the generalized category discovery (GCD) \cite{vaze2022generalized}, where unlabeled images are from both novel and labeled categories. 

\par
GCD is a challenging open-world problem in that we need to 1) jointly distinguish the known and unknown classes and 2) discover the novel clusters without any annotations. To solve this problem, Vaze \etal \cite{vaze2022generalized} leverage the contrastive learning technique to learn discriminative representation for unlabeled data and use $k$-means~\cite{macqueen1967classification} to obtain final clustering results. In this method, the labeled data are fully exploited by supervised contrastive learning. However, self-supervised learning is applied to the unlabeled data, which enforces samples to be close to their augmentation counterparts while far away from others. As a consequence, the underlying relationships between samples of the same conceptions are largely overlooked and thus will lead to degraded representation learning. Intuitively, samples that belong to the same conceptions should be similar to each other in the feature space. The conceptions can be regarded as: classes, super-classes, sub-classes, etc. For example, as shown in ~\cref{fig:conceptual}, samples of the same class should be similar to each other, \textit{e.g.,} samples of the bus, samples of the bicycle. In addition, in the super-classes view, classes of the transportation, \textit{e.g.}, Bus and Bicycle, should belong to the same concept. Hence, the samples of transportation should be closer than that of other concepts (\textit{e.g.}, animals). Similarly, samples belong to the same sub-classes (\textit{e.g.,} red train) should be closer to that of other sub-classes (\textit{e.g.,} white train). Hence, embracing such conceptions and their relationships can greatly benefit the representation learning for unlabeled data, especially for unseen classes.
\par
Motivated by this, we propose a Dynamic Conceptional Contrastive Learning (DCCL) framework for GCD to effectively leverage the underlying relationships between unlabeled data for representation learning. Specifically, our DCCL includes two steps: Dynamic Conception Generation (DCG) and  Dual-level Contrastive Learning (DCL). In DCG, we dynamically generate conceptions based on the hyper-parameter-free clustering method equipped with the proposed semi-supervised conceptional consolidation. In DCL, we propose to optimize the model with conception-level and instance-level contrastive learning objectives, where we maintain a dynamic memory to ensure comparing with the up-to-date conceptions. The DCG and DCL are alternately performed until the model converges.
\par
We summarize the contributions of this work as follows:
\begin{itemize}
    \item We propose a novel dynamic conceptional contrastive learning (DCCL) framework to effectively leverage the underlying relationships between unlabeled samples for learning discriminative representation for GCD.
    
    \item We introduce a novel dynamic conception generation and update mechanism to ensure consistent conception learning, which encourages the model to produce more discriminative representation.
    
    \item Our DCCL approach consistently achieves superior performance over state-of-the-art GCD algorithms on both generic and fine-grained tasks.
\end{itemize}

\begin{figure*}[!t]
  \begin{center}
  \includegraphics[width=0.98\textwidth]{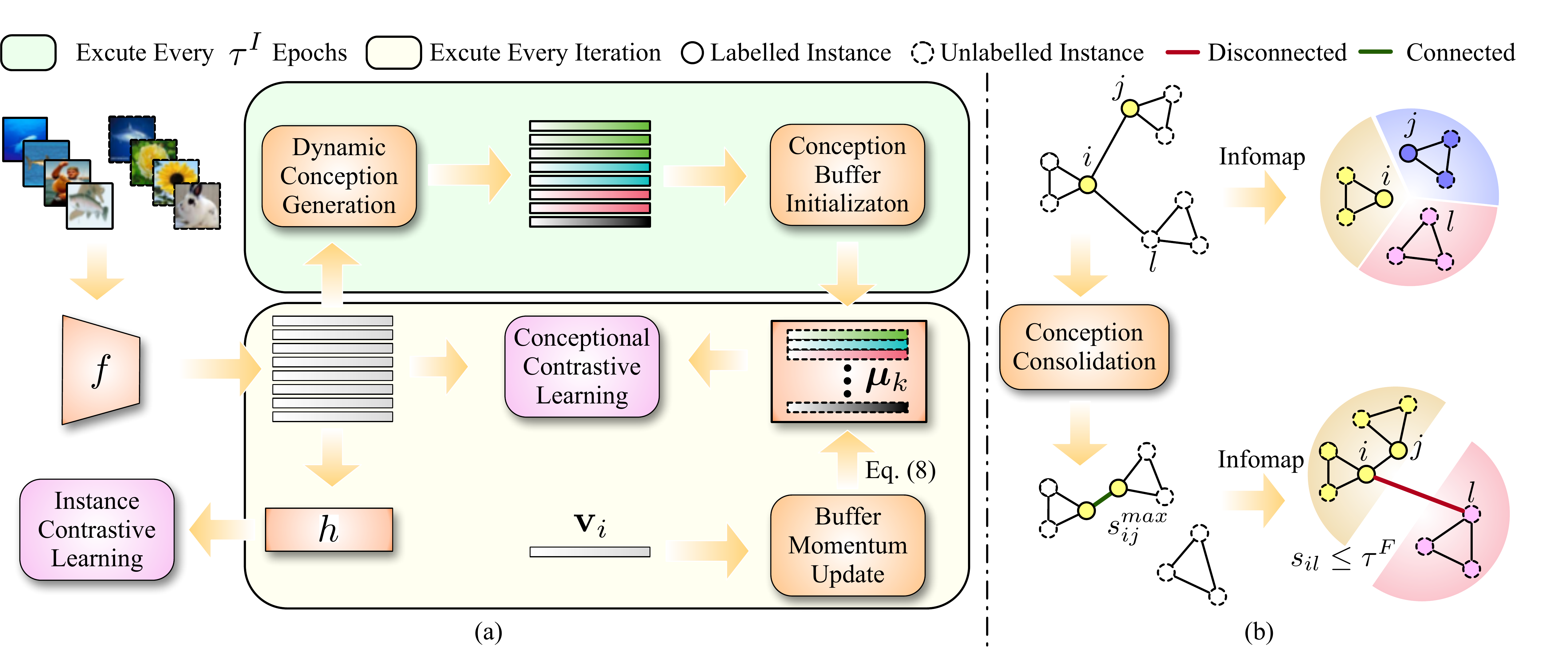}
\caption{(a) Overview of our DCCL framework. We first extract features and cluster the features to generate conceptional labels, then initialize conception representations by our DCG, and last learn representations by joint instance-level and conception-level objectives. During the training process, the DCG and dual-level representation learning are performed alternately, in which the conception buffer is updated every iteration to keep the consistency of the changing instance features and conceptional representations. (b) Illustration of the proposed conception consolidation. Without consolidating the relationships of conceptions by label information, Infomap tends to over-cluster data and thus provides the supervision that has a high risk to over-correct affinities between neighbor instances.
\label{fig:framework}}
  \end{center}
\vspace{-2em}
\end{figure*}

\section{Related Work}
\subsection{Novel Category Discovery} 
\par\noindent
Novel category discovery~\cite{han2019learning} (NCD) tasks aim at discovering new categories by leveraging the knowledge of a set of labeled categories. RankStat~\cite{han2021autonovel} indicates that self-supervised pre-training is helpful for NCD. NCL~\cite{zhong2021neighborhood} adopts contrastive learning to improve representation learning. UNO~\cite{fini2021unified} proposes a unified objective for jointly learning on unlabeled and labeled data. Most recently, NCD has been extended to a generalized category discovery (GCD)~\cite{vaze2022generalized}, in which the unlabeled data include both labeled and unlabeled categories. Later, ORCA~\cite{DBLP:conf/iclr/CaoBL22} defines an open-world semi-supervised learning task, which is similar to GCD. Although these definitions are relatively practical, most methods remain to assume that the class number of clustered data is known. Nevertheless, such prior knowledge is often not acquired in advance for real-world applications. To handle the drawback, DTC~\cite{han2019learning} and GCD~\cite{vaze2022generalized} employed an independent algorithm to search the optimal class number, however, they did not associate clustering estimation with representation learning. \textit{Unlike these offline algorithms, we propose to jointly consider downstream clustering and representation learning. The experimental results show they are mutually beneficial for each other.}
\subsection{Contrastive Learning based on Memory Buffer}
\par\noindent
Contrastive learning~\cite{chen2020simple,chen2020simclr,noroozi2016contrastive,wang2021,wang2022} (CL) has been shown to be significantly effective for representation learning in a self-supervised manner. MoCo~\cite{he2020momentum} demonstrates that sampling positive-negative pairs from an instance-level buffer can benefit CL and reduce the impact of the size of the training batch. Then, instead of contrasting over all instances in a mini-batch, prototypical contrastive learning~\cite{DBLP:conf/iclr/0001ZXH21} (PCL) that contrasts the instance features with a set of prototypes, has been shown to provide comprehensive supervision. However, PCL still needs an instance-level memory buffer to yield the prototype set, which is not computation- and memory-efficient. Recently, SCL~\cite{DBLP:conf/bmvc/HuangG22} propose a cluster-level momentum encoder but considers only three fixed numbers of classes during training, which is still limited compared to our dynamic method. \textit{Different from PCL and SCL that consider the fixed numbers of classes during the whole training process, our DCCL dynamically estimates the number of classes for different training stages in a efficient way, which encourages models to learn more discriminative representation.}

\subsection{Semi-Supervised Learning and Clustering}
\par\noindent
Semi-supervised learning (SSL) has been a long-standing research topic~\cite{yang2021survey}. Different from GCD, SSL often assumes that the labeled and unlabeled data come from the same set of classes, in which consistency-based methods are the most effective methods for SSL, such as Mean-teacher \cite{ tarvainen2017mean}, MixMatch~\cite{berthelot2019mixmatch}, and FixMatch\cite{sohn2020fixmatch}. Consistency-based methods are the most effective methods for SSL, such as Mean-teacher \cite{ tarvainen2017mean}, MixMatch~\cite{berthelot2019mixmatch}, and FixMatch\cite{sohn2020fixmatch}. Moreover, semi-supervised classification is a relatively well-defined task, while the supervised information in semi-supervised clustering can take different forms~\cite{lange2005learning}, such as two instances are known to be must-linked in a relationship matrix or some cluster assignments are known beforehand. For instance, Basu \textit{et al.}~\cite{DBLP:conf/icml/BasuBM02} proposed to initialize the clusters based on the data points for which cluster assignments are known. However, these methods can not adaptively assign the number of clusters for generating conceptions. \textit{To mitigate these limitations, we improve the classical InfoMap~\cite{rosvall2008maps} to fully leverage both labeled and unlabeled data and dynamically produce varying conceptional labels for different contrastive learning epochs.}

\section{Dynamic Conceptional Contrastive Learning}

\subsection{Problem Formulation}
Generalized category discovery (GCD) aims at automatically categorizing unlabeled images in a dataset, where the partial data are labeled and the remaining are unlabeled. The unlabeled images come from either labeled (known) classes or unlabeled (unknown) classes. This is a more realistic open-world setting than the common closed-set classification that assumes the labeled and unlabeled data belong to the same classes. Let the dataset be $\mathcal{D} = \mathcal{D}^{L} \cup  \mathcal{D}^{U} $, where $\mathcal{D}^{L}=\{(\mathbf{x}^{L}_{i}, \mathbf{y}^{L}_{i})\}^{M^{L}}_{i=1} \in \mathcal{X} \times\mathcal{Y}^{L}$, $L$ denotes the labeled subset and $\mathcal{D}^{U}=\{(\mathbf{x}^{U}_{i}, \mathbf{y}^{U}_{i})\}^{M^{U}}_{i=1} \in \mathcal{X}\times \mathcal{Y}^{U}$ denotes the unlabeled subset with
unknown $\mathbf{y}^{U}_{i}\in \mathcal{Y}^{U}$. Only a subset of classes contains labeled instances, i.e., $\mathcal{Y}^{L}\subset \mathcal{Y}^{U}$. The number
of labeled classes $N^{L}$ can be directly calculated from the labeled data, while the number of unlabeled classes $N^{U}$ is not known during model training. Let $f$ and $h$ be a feature extractor and a MLP projection head. The extracted representation is defined as $\mathbf{v}_{i} = f(\mathbf{x}_{i})$.

\subsection{Overview}
\par
To tackle the problem of GCD, we propose a novel framework DCCL (see Fig.~\ref{fig:framework}), to jointly learn representations using dual-level contrastive learning (DCL) and explore all possible relationships between labeled and unlabeled instances in a conceptional view. First, we extend the classical unsupervised clustering algorithm, Infomap~\cite{rosvall2008maps}, to a semi-supervised manner, which aims at dynamically generating reasonable conceptional representations and associating the labeled and unlabeled instances during representation learning. By alternately executing the dynamic conception generation (DCG) and DCL, DCL benefits from informative supervised information to generate higher-quality representations. Meanwhile, DCG gradually produces more comprehensive guidance based on a deeper understanding of conceptual relationships. DCG and DCL mutually benefit each other, thus resulting in a better representation for downstream clustering.

\subsection{Dynamic Conception Generation}\label{sec:label_generation}
Although the test-time semi-supervised $k$-means~\cite{vaze2022generalized} (SSK) succeeds in achieving considerable performance gain, it fails to jointly consider the representation learning and supervision information from semi-supervised clustering. Moreover, it is infeasible to directly perform SSK in each training epoch for assigning pseudo labels, because the real number of clusters is unknown during training. To overcome these drawbacks, we propose a dynamic conception generation (DCG) based on the hyper-parameter-free Infomap~\cite{rosvall2008maps} algorithm. Specifically, we first propose a semi-supervised conceptional consolidation method to construct a similarity network, then execute the Infomap clustering algorithm on the constructed network to get conceptional label assignments, and finally calculate conception representation and initialize conceptional memory buffer.
\par\noindent
\textbf{Conception Consolidation.} In a given network, Infomap aims at partitioning semantic-similar sub-networks by the pattern of connections. To leverage the supervision from labeled data, we propose to enforce the similarity constraints into the networking, according to the labeled data that belong to the same category. Formally, we construct an adjacent matrix $\mathcal{A}$ to represent the possible connection relationships among all instances. The weight of the edge of the $i$-th and $j$-th instances is given by:
\begin{equation}\label{eq:adjecent}
    \mathcal{A}_{ij}= \begin{cases}s^{max}_{i}, & \text { if } \mathbf{y}_{i}, \mathbf{y}_{j} \in \mathcal{Y}^{L} \text{ and } \mathbf{y}_{i}=\mathbf{y}_{j}\\ 
    s_{ij}, & \text { if }\mathbf{y}_{i} \text{ or } \mathbf{y}_{j} \in \mathcal{Y}^{U} \text{ and } s_{ij} \textgreater \tau^{F}\\
    0, & \text { otherwise }\end{cases}
\end{equation}
\begin{equation}\label{eq:max_s}
    s^{max}_{i}=\underset{j}{\arg \max }\left\{s_{ij} \mid j \in \mathcal{D}\right\},
\end{equation}
\begin{equation}\label{eq:s_ij}
    s_{ij}=[(\mathbf{v}_{i}/{\left \| \mathbf{v}_{i} \right \| })\bm{\cdot} (\mathbf{v}_{j}/\left \| \mathbf{v}_{j} \right \|) + 1]/2\in [0,1],
\end{equation}
where $\bm{\cdot}$ denotes dot product and $\left \|\cdot \right \|$ is $l_{2}$ normalization. The $\tau^{F}$ is a threshold to select high-confidence links. Through the conception consolidation illustrated in~\cref{fig:framework}(b), we can establish a reliable relationship network with rich structural information for the subsequent clustering.
\par\noindent
\textbf{Remark.} We set the similarities of positive pairs with the maximal value of neighborhood similarities instead of 1. This is because we experimentally find that when imposing 1 for constraining positive pairs, Infomap tends to group all the labeled positive instances as an individual cluster.
\par\noindent
\textbf{Entropy Minimization Clustering.} In Infomap algorithm~\cite{rosvall2009map}, the clustering problem is equivalent to minimizing the entropy that represents the minimum description length of the coding network. By solving the minimization objective, we acquire a conceptional label set $\mathcal{C} = \{\mathbf{c}_{i}\}^{M^{L}+M^{U}}_{i=1} \in \mathcal{Y}^{G}$ for both labeled and unlabeled instances. Then, we combine the extracted feature vectors $\mathcal{V}$ and corresponding conceptional labels to construct a generated feature dataset $\mathcal{D}^{G} = \{(\mathbf{v}_{i}, \mathbf{c}_{i})\}^{M^{L}+M^{U}}_{i=1} \in \mathcal{V}\times \mathcal{Y}^{G}$. $|\mathcal{Y}^{G}|$ denotes the number of estimated conceptions. 
\par\noindent
\textbf{Conceptional Memory Initialization.}
In this paper, our DCCL maintains a conception-level memory buffer that provides dynamic conceptional representations for dual-level contrastive learning, which is elaborated in \cref{sec:two_level}. Here, we introduce the initialization of the conceptional memory buffer (CMB). We use the mean feature vector of the instances that share the same conceptional label to form a unique conceptional representation.
Formally, the initial conceptional representation set is defined as:
\begin{equation}\label{eq:initialization}
\mathcal{U} = \{\bm{\mu}_{k}\}^{K}_{k=1},~~~\bm{\mu}_{k} = \frac{1}{|\mathcal{D}^{G}_{k}|}\sum_{\mathbf{v}_{i}\in \mathcal{D}^{G}_{k} }\mathbf{v}_{i},~~~K = |\mathcal{Y}^{G}|,
\end{equation}
where $\mathcal{D}^{G}_{k}$ denotes the $k$-th conception subset, \ie, if $\mathbf{v}_{i}\in \mathcal{D}^{G}_{k}, \mathbf{c}_{i} = k$. During whole training process, the initialization of CMB is executed every $\tau^{I}$ epochs on center-cropped images. Thus, the number of conceptions $K$ is dynamically changing along with model training.

\begin{algorithm}[t]
\SetAlgoLined

\KwIn{Feature Extractor $f$, Projection Head $h$, Labeled data $\mathcal{D}^{L}$ and Unlabeled data $\mathcal{D}^{U}$.}
\KwOut{$f$ and $h$.}
\For{$n=1$ \textbf{in} $[1, max\_epoch]$}{
\If{$n$ \textbf{mod} $\tau^{I} == 0$}{
Extract features and construct adjacency matrix $\mathcal{A}$ by~\cref{eq:adjecent}, \cref{eq:max_s} and \cref{eq:s_ij}\;
Perform InfoMap~\cite{rosvall2008maps} clustering to assign conceptional labels $\mathcal{C}$\;
Initialize conceptional buffer by \cref{eq:initialization}\;
}
{
\For{$i=1$ \textbf{in} $[1, max\_iteration]$}{
Sample mini-batches from $\mathcal{D}^{L} \cup\mathcal{D}^{U}$\;
Calculate overall optimization objective by \cref{eq:total_loss}\;
Update $f$ and $h$ by SGD~\cite{qian1999momentum}\;
Update conceptional buffer by~\cref{eq:update}\;
}

}
}

\caption{Algorithm Pipeline of our DCCL\label{algorithm:DCCL}}

\end{algorithm}

\subsection{Dual-Level Contrastive Learning}\label{sec:two_level}
In this section, we first explain the proposed conception-level contrastive learning, then elaborate on the update of the conceptional memory buffer, and finally introduce the employed instance-level contrastive learning approach. 
\par\noindent
\textbf{Conception-Level Contrastive Learning.}
Based on the generated conceptional representations in \cref{sec:label_generation}, we propose to perform conception-level contrastive learning. Specifically, we first sample $N^{C}$ conception labels and a fixed number $N^{I}$ of instances for each conception label, resulting in a mini-batch $\mathcal{B}^{C}$ with $N^{C}\times N^{I}$ instances. Next, each instance representation is compared to all the conceptional representations. We pull the instance representation from its corresponding conceptional representation closer and push the instance representation away from other conceptional representations. Formally, we define the conceptional contrastive loss function as the following:

\begin{equation}
    \mathcal{L}^{C}_{i}=-\log \frac{\exp \left(\mathbf{v}_i \bm{\cdot} \bm{\mu}_{\mathbf{c}_{i}} / \tau^{C}\right)}{\sum_{k=1, k\neq \mathbf{c}_{i}}^{K} \exp( \mathbf{v}_{i} \bm{\cdot} \bm{\mu}_{k}/ \tau^{C})},
\end{equation}
where $\tau^{C}$ is a temperature hyper-parameter to control the strength of the conception-level contrastive learning. In addition, in order to explicitly encourage learned representations with a large inter-conception margin, we propose a dispersion loss to further push the different conception representations away from each other. The loss function for the $m$-th and the $n$-th conceptions in $\mathcal{B}^{C}$ is:
\begin{equation}
    \mathcal{L}(m,n)= \Big[||\frac{1}{|\mathcal{B}^{C}_{m}|}\sum_{\mathbf{v}_{i}\in \mathcal{B}^{C}_{m} }\mathbf{v}_{i}||\bm{\cdot}||\frac{1}{|\mathcal{B}^{C}_{n}|}\sum_{\mathbf{v}_{j}\in \mathcal{B}^{C}_{n} }\mathbf{v}_{j}||-\tau^{M}\Big]_{+},
\end{equation}
where $\tau^{M}$ is a threshold to filter the conception pairs with high uncertainty. We assume that two conception representations that are close tend to be highly entangled conceptions. Separating these conceptions has a high risk of over-correct. We explore the impact of $\tau^{M}$ in~\cref{sec:hyper_parameter}. The dispersion loss function over a mini-batch is defined as:
\begin{equation}
    \mathcal{L}^{D}= \frac{1}{N^{C}}\sum^{N^{C}}_{m=1} \frac{1}{N^{C}}\sum^{N^{C}}_{n=1}\mathcal{L}(m,n).
\end{equation}
\par\noindent
\textbf{Conceptional Memory Update.}
Different from \cite{he2020momentum,DBLP:conf/iclr/0001ZXH21} that need to save all training instances, our CMB stores only the conception representations, which significantly reduces storage cost. Furthermore, the instance-wise update is easy to lead to an inconsistent update during each training iteration. To mitigate this drawback, we first adopt the re-sampling method as detailed in the previous section. Next, we propose to update the conception representation by each corresponding instance feature following a momentum update mechanism. The process is formulated as:
\begin{equation}\label{eq:update}
\bm{\mu}_{\mathbf{c}_{i}} \leftarrow \eta \bm{\mu}_{\mathbf{c}_{i}}+(1-\eta)\mathbf{v}_{i},
\end{equation}
where $\eta$ is the momentum updating factor.
\par\noindent
\textbf{Remark.} DCCL updates the memory buffer and computes the losses both at the conceptional level, which consistently updates the conceptional representation to maintain the conceptional consistency during the whole training process.

\par\noindent
\textbf{Instance-Level Contrastive Learning.}
Inspired by \cite{vaze2022generalized}, we combine supervised contrastive loss and self-supervised contrastive loss as an instance contrastive loss (ICL), to fine-tune the model. Formally, we assume $x_{i}$ and $\hat{x}_{i}$ are two views (random augmentations) of the same image in a randomly-sampled mini-batch $\mathcal{B}^{I}$. Let $h$ be a MLP projection head. The extracted representation $\mathbf{v}_{i}$ is further projected by $h$ to high-dimensional embedding space for instance-level contrastive learning. The loss function is: 
\begin{equation}
\begin{aligned}
&\mathcal{L}^{I}_{i}=(\lambda-1)\log \frac{\exp \left(h\left(\mathbf{v}_i\right) \cdot h\left(\hat{\mathbf{v}}_i\right) / \tau\right)}{\sum_{j\in \mathcal{B}^{I}, j \neq i} \exp \left(h\left(\mathbf{v}_i\right) \cdot h\left(\mathbf{v}_j\right) / \tau^{S}\right)}\\
&\frac{-\lambda}{|\mathcal{P}(i)|} \sum_{p \in \mathcal{P}(i)} \log \frac{\exp \left(h\left(\mathbf{v}^{L}_i\right) \cdot h\left(\mathbf{v}^{L}_p\right) / \tau\right)}{\sum_{j\in \mathcal{B}^{L}, j \neq i} \exp \left(h\left(\mathbf{v}^{L}_i\right) \cdot h\left(\mathbf{v}^{L}_j\right) / \tau^{L}\right)},\\
\end{aligned}
\end{equation}
where $\mathcal{B}^{L}$ donates the labeled subset within the mini-batch $\mathcal{B}^{I}$ and $\mathcal{B}^{I} =  \mathcal{B}^{L} \cup \mathcal{B}^{U}$. $\mathcal{P}(i)$ is the positive index set for the anchor image $i\in \mathcal{B}^{L}$. $\lambda$ is a trade-off factor to balance the contributions of self-supervised and supervised learning. For a fair comparison, we follow \cite{vaze2022generalized} and set $\lambda$ to 0.35.

\begin{table*}[th]
\centering
\caption{Statistics of the datasets and the splits for GCD. The first three are generic datasets while the last three are fine-grained datasets.
\label{tab:dataset}}
\resizebox{0.8\textwidth}{!}{
\begin{tabular}{cccccccc}
\cmidrule[1pt]{1-8}
\multicolumn{2}{c}{Dataset}             & CIFAR10~\cite{krizhevsky2009learning} & CIFAR100~\cite{krizhevsky2009learning} & ImageNet-100~\cite{deng2009imagenet} & CUB-200~\cite{wah2011caltech} & SCars~\cite{krause20133d} & Pet~\cite{parkhi2012cats}\\ \cmidrule[1pt]{1-8}
\multirow{2}{*}{Labelled}   & \#~Classes & 5       & 80       & 50           & 100     & 98             & 19    \\
                            & \#~Images  & 12,500  & 20,000   & 31,860       & 1,498   & 2,000 & 942   \\ \cmidrule[0.5pt]{1-8}
\multirow{2}{*}{Unlabelled} & \#~Classes & 10      & 100      & 100          & 200     & 196            & 37    \\
                            & \#~Images  & 37,500  & 30,000   & 95,255       & 4496    & 6,144 & 2,738 \\ \cmidrule[1pt]{1-8}
\end{tabular}
} 
\vspace{-1em}
\end{table*}
\subsection{Joint Optimization}
During the whole training process, we alternately perform dynamic conception generalization and dual-level contrastive representation learning, until the maximal training epoch. The pseudo-code of DCCL is elaborated in~\cref{algorithm:DCCL}. The overall objective over on mini-batch $\mathcal{B}$ is given by the weighted sum of each loss function:
\begin{equation}\label{eq:total_loss}
    \mathcal{L}_{total} = \frac{1}{|\mathcal{B}|}\sum_{i\in \mathcal{B}}\mathcal{L}^{I}_{i} + \alpha\frac{1}{|\mathcal{B}|}\sum_{i\in \mathcal{B}}\mathcal{L}^{C}_{i} + \beta\mathcal{L}^{D},
\end{equation}
where $ \alpha$ and $\beta$ are the weights to adjust the strengths of two loss functions. In all experiments, we use $l_{2}$ normalized feature vector $||\mathbf{v}||_{2}$ for clustering evaluation.

\section{Experiments}

\subsection{Experimental Setup}

\begin{table}[t]
\centering
\caption{Results on generic image recognition datasets.\label{tab:generic}}
\resizebox{0.48\textwidth}{!}{%
\begin{tabular}{lccccccccc}
\cmidrule[1pt]{1-10}
\multicolumn{1}{c}{\multirow{2}{*}{Method}} & \multicolumn{3}{c}{CIFAR10} & \multicolumn{3}{c}{CIFAR100} & \multicolumn{3}{c}{ImageNet-100} \\ \cmidrule[0.5pt]{2-10}
\multicolumn{1}{c}{}                        & All      & Old     & New     & All      & Old     & New     & All       & Old       & New      \\ \cmidrule[0.5pt]{1-10}
k-means                                     & 83.6     & 85.7    & 82.5    & 52.0     & 52.2    & 50.8    & 72.7      & 75.5      & 71.3     \\
RankStats+                                  & 46.8     & 19.2    & 60.5    & 58.2     & 77.6    & 19.3    & 37.1      & 61.6      & 24.8     \\
UNO+                                        & 68.6     & \textbf{98.3}    & 53.8    & 69.5     & 80.6    & 47.2    & 70.3      & 95.0      & 57.9     \\
GCD & 91.5     & 97.9    & 88.2    & 73.0     & 76.2    & 66.5    & 74.1      & 89.8      & 66.3     \\ \hline
DCCL                                         & \textbf{96.3}         &   96.5     &   \textbf{96.9}      &	\textbf{75.3}		
      &    \textbf{76.8}     & \textbf{70.2}   &  	\textbf{80.5}&   \textbf{90.5}     &   \textbf{76.2}       \\ \cmidrule[1pt]{1-10}
\end{tabular}
}
\vspace{-1em}
\end{table}

\par\noindent
\textbf{Data and Evaluation Metric.} We evaluate DCCL on three generic image classification datasets, namely CIFAR-10~\cite{krizhevsky2009learning}, CIFAR-100~\cite{krizhevsky2009learning} and ImageNet-100~\cite{vaze2022generalized}. ImageNet-100 denotes randomly sub-sampling 100 classes from the ImageNet~\cite{deng2009imagenet} dataset. The dataset statistics and train-test splits are described in~\cref{tab:dataset}. We further evaluate DCCL on three more challenging fine-grained image classification datasets: CUB-200~\cite{wah2011caltech}, Stanford Cars~\cite{krause20133d}, and Oxford-IIIT Pet~\cite{parkhi2012cats}. The original training set of each fine-grained dataset is separated into labeled and unlabeled parts. We follow~\cite{vaze2022generalized} sample a subset of half the classes as ``Old'' categories. 50\% of instances of each labeled class are drawn to form the labeled set, and all the remaining data constitute the unlabeled set. For evaluation, we measure the clustering accuracy by comparing the predicted label assignment with the ground truth, following the protocol in~\cite{vaze2022generalized}.
\par\noindent
\textbf{Implementation Details.} We adopt the ViT-B-16 pre-trained by DINO~\cite{caron2021emerging} as our backbone network. The output [CLS] token is used as the feature representation, which is also used for conception-level contrastive learning. Following~\cite{vaze2022generalized}, we project the representations by a projection head and use the projected embeddings for instance-level contrastive learning. We set the dimension of projected embeddings to 65,536 following~\cite{caron2021emerging}. At training time, we feed two views with random augmentations to the model. We only fine-tune the last block of the vision transformer with an initial learning rate of 0.01 and the head is trained with an initial learning rate of 0.1. All methods are trained for 200 epochs with a cosine annealing schedule. The size of the mini-batch is set to 128 with $N^{C}$=8 and $N^{I}$=16. For a fair comparison, we follow~\cite{vaze2022generalized} and set the temperatures of two supervised contrastive losses $\tau^{S}$, $\tau^{L}$ and $\tau^{C}$ to 0.07, 0.05 and 0.05, respectively. The $\tau^{F}$ is empirically set to 0.6 for the fine-grained datasets and 0.7 for the generic datasets. Other hyper-parameters are discussed in~\cref{sec:hyper_parameter}. In testing, we first use the class number estimation algorithm~\cite{vaze2022generalized} to predict the number of classes of the testing dataset, and then use semi-supervised $k$-means for clustering evaluation. In dynamic conception generation, we adopt faiss \cite{johnson2019billion} to accelerate the construction of relationship networks. Our experiments are conducted on RTX 3090 GPUs.
\subsection{Comparison with State-of-the-Art}
To evaluate the performances of our DCCL, we conduct three group experiments by comparing our DCCL with three strong GCD baselines, including RankStats~\cite{han2021autonovel} and UNO~\cite{fini2021unified} and the state-of-the-art GCD method~\cite{vaze2022generalized}.
\par\noindent
\textbf{Comparison on Generic Datasets.}
As shown in~\cref{tab:generic}, our DCCL is compared with other competitors on the generic image recognition datasets. Overall, the results in~\cref{tab:generic} show our DCCL consistently outperforms all others by a significant margin. Specifically, DCCL outperforms the GCD method~\cite{vaze2022generalized} by 4.8\% on CIFAR-10, 2.3\% on CIFAR-100, and 6.4\% on ImageNet-100 for ‘All’ classes, and by 8.7\% on CIFAR-10, 3.7\% on CIFAR-100, and 9.9\% on ImageNet-100 for ‘Unseen’ classes. These results experimentally demonstrate the generated dynamic conceptions provide effective supervision to learn better representations for unlabeled data. Moreover, UNO+ shows a strong accuracy on ``Old'' classes, but its accuracy when testing on ``New'' classes is relatively lower. This is because UNO+ trains the linear classifier on ``Old'' classes, thus resulting in an inevitable bias. On the contrary, our DCCL gets a relatively good balance on both the ``Old'' and ``Unseen'' classes, without bias to the labeled data.

\par\noindent
\textbf{Comparison on Fine-Grained Datasets.}
In general, the differences between different classes in fine-grained datasets are subtle, which leads the fine-grained visual understanding to be more challenging for GCD. For verifying the effects of DCCL on fine-grained tasks, we compare our method with others on fine-grained image recognition datasets. The results in~\cref{tab:fine-grained} show that DCCL consistently outperforms all other methods for ``All'' and ``New'' classes. Specifically, on CUB-200 and Oxford-Pet, DCCL achieves 12.2\% and 7.9\% improvement over the state-of-the-art for ``All'' classes. Especially for ``New'' classes, DCCL outperforms GCD by 16.2\% on the CUB-200 dataset. These results demonstrate that our DCCL is efficient in capturing the conceptional information shared across different fine-grained classes, thereby generating precise and helpful supervision for representation learning.

\begin{table}[t]
\centering
\caption{Results on fine-grained datasets.\label{tab:fine-grained}}
\resizebox{0.48\textwidth}{!}{%
\begin{tabular}{lccccccccc}
\cmidrule[1pt]{1-10}
\multicolumn{1}{c}{\multirow{2}{*}{method}} & \multicolumn{3}{c}{CUB-200} & \multicolumn{3}{c}{Stanford-Cars} & \multicolumn{3}{c}{Oxford-Pet} \\ \cmidrule[0.5pt]{2-10}
\multicolumn{1}{c}{}                        & All      & Old     & New     & All      & Old     & New     & All       & Old       & New         \\ \cmidrule[0.5pt]{1-10}
k-means                                     & 34.3 &38.9 &32.1 &12.8& 10.6& 13.8& 77.1& 70.1& 80.7     \\
RankStats+  & 33.3 &51.6& 24.2& 28.3& 61.8& 12.1&
-&-&-\\
UNO+ &35.1& 49.0& 28.1& 35.5&\textbf{70.5}&18.6&-&- &-    \\ 
GCD&51.3&56.6&
48.7 &39.0& 57.6& 29.9& 80.2 &85.1& 77.6 \\
\hline

DCCL                                         &\textbf{63.5}          &    \textbf{60.8}     &  \textbf{64.9}       &     \textbf{43.1}   & 55.7 &   \textbf{36.2}         &      \textbf{88.1}   &   \textbf{88.2}        &   \textbf{88.0}
       \\ \cmidrule[1pt]{1-10}
\end{tabular}
}
\vspace{-1.5em}
\end{table}

\begin{figure*}[t]
  \begin{center}
  \includegraphics[width=0.98\textwidth]{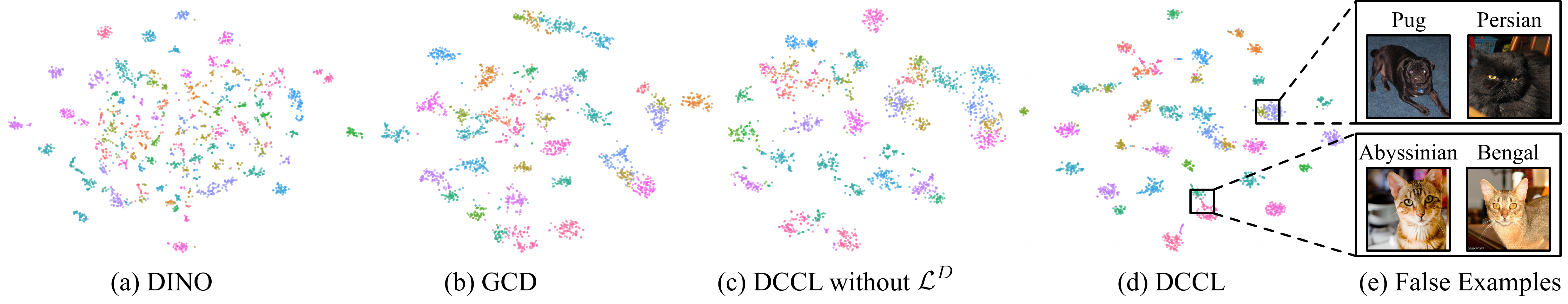}
\caption{Visualization of features distributions of the unlabeled set of the Pet~\cite{parkhi2012cats} dataset. (a)-(d) are the results generated from DINO~\cite{caron2021emerging}, GCD~\cite{vaze2022generalized}, our DCCL without $\mathcal{L}^{D}$ and full DCCL, in turn. (e) is a visualization of false samples that are easy to be incorrectly clustered.
\label{fig:distribution}}
  \end{center}
\vspace{-2em}
\end{figure*}

\par\noindent
\textbf{Visualization of Feature Distributions.}
To qualitatively explore the clustered features on Pets dataset~\cite{parkhi2012cats}, we visualize the t-SNE embeddings projected from the features extracted by pre-trained ViT~\cite{caron2021emerging}, GCD~\cite{vaze2022generalized}, the DCCL without the proposed dispersion loss and our full DCCL method. As shown in~\cref{fig:distribution}, our features are more discriminative than the features from the pre-trained ViT and GCD. By comparing~\cref{fig:distribution}(c) and~\cref{fig:distribution}(d), the proposed dispersion loss effectively pushes cluster centers away from each other. A large inter-cluster margin not only improves cluster boundaries for ``Old'' and ``New'' categories, but also compacts intra-cluster distribution.
\par\noindent
\textbf{Summary.} The experimental results show that our DCCL achieves significant improvements on both generic and fine-grained datasets. Especially for discovering ``New'' categories in challenging fine-grained tasks, our dynamic conceptional contrastive learning succeeds in mining shared conceptions, which is especially beneficial for the generalized fine-grained new category discovery.

\begin{table}[!t]
\centering
\caption{Effectiveness of each component of our DCCL. ``MU'' and ``CC'' denote the proposed momentum update by~\cref{eq:update} and the conception consolidation proposed in~\cref{sec:label_generation}, respectively. \label{tab:ablation}}
\resizebox{0.48\textwidth}{!}{
\begin{tabular}{c|ccccc|cccccc}
\cmidrule[1pt]{1-12}
\multirow{2}{*}{Index}&\multicolumn{5}{c|}{Component}                                   & \multicolumn{3}{c}{CUB-200~\cite{wah2011caltech}} & \multicolumn{3}{c}{Pet~\cite{parkhi2012cats}} \\ \cmidrule[0.5pt]{2-12}
 &$\mathcal{L}^{I}$ & $\mathcal{L}^{C}$ & $\mathcal{L}^{D}$ &MU & CC & All          & Old          & New         & All         & Old        & New        \\ \cmidrule[1pt]{1-12}
a)&\cmark  &       &       &                 &                          & 51.3         & 56.6         & 48.7        &     80.2    & 85.1& 77.6 \\

b)& &  \cmark &   & & & 54.9         & 52.3         & 55.4        & 81.6        & 80.7       & 81.0       \\
c)& & \cmark&       &  \cmark  &                          & 57.7         & 54.0         & 58.1        & 83.5       & 81.1       & 80.3       \\
d)& &  \cmark & \cmark &  \cmark   &                          & 59.5         & 53.3         & 60.8        & 84.3        & 83.1       & 84.5       \\
e) & &  \cmark & \cmark &  \cmark   &    \cmark                      & 60.1 & 59.4 & 60.7& 85.8& 86.8 & 84.6 \\
f)&\cmark  & \cmark &  \cmark &   \cmark    &  \cmark                &     \textbf{63.5}         & \textbf{60.8 }            &\textbf{64.9}             & \textbf{88.1}  & \textbf{88.2} & \textbf{88.0} \\ \cmidrule[1pt]{1-12}
\end{tabular}
}
\vspace{-1em}
\end{table}

\subsection{Effectiveness Evaluation}\label{sec:ab}
To verify the effectiveness of each component in our DCCL, we conduct five experiments on both CUB-200~\cite{wah2011caltech} and Pet~\cite{parkhi2012cats} datasets, as shown in~\cref{tab:ablation}. Note that the configuration of the experiment a) is the same with GCD~\cite{vaze2022generalized}, which is the baseline method in our experiments.
\par\noindent
\textbf{Effectiveness of Conceptional contrastive Learning.}
Based on the results of the experiment a) and b), we find that using only the conceptional contrastive learning can achieves competitive performance, compared to baseline method with only instance-level contrastive learning. 
\par\noindent
\textbf{Effectiveness of Conception-Level Momentum Updating.} Comparing the experiment b) and c), we can find that consistent update of conceptional representations by the proposed momentum update can bring considerable improvements on both new and old classes. This implies that due to periodically generating conceptional representations, the conceptional labels are kept fixed within one training period, which leads to a severe sub-optimal problem, during conceptional contrastive learning. Thus, the proposed momentum update mitigates the problem to some extent. 
\par\noindent
\textbf{Effectiveness of Dispersion Loss.} The experiment d) in~\cref{tab:ablation} shows that by adding the proposed dispersion loss, the model's performances on new classes acquire further improvements by 2.7\% on ``New'' classes for the CUB-200 dataset. The improvements can be observed in~\cref{fig:distribution}(d).
\par\noindent
\textbf{Effectiveness of Conception Consolidation.}
Without the proposed conception consolidation that considers labeled information to impose semi-supervised constraints, our DCCL suffers from a performance balance between the new and the old classes, as shown in~\cref{tab:ablation}~b), c) and d). In the experiment e) and f), our full method shows superior performance on all evaluation metrics, which experimentally demonstrates that our conception consolidation plays an essential role in rectifying the estimated latent conception relations between seen and unseen classes.

\begin{figure}[!t]
  \begin{center}
  \includegraphics[width=0.48\textwidth]{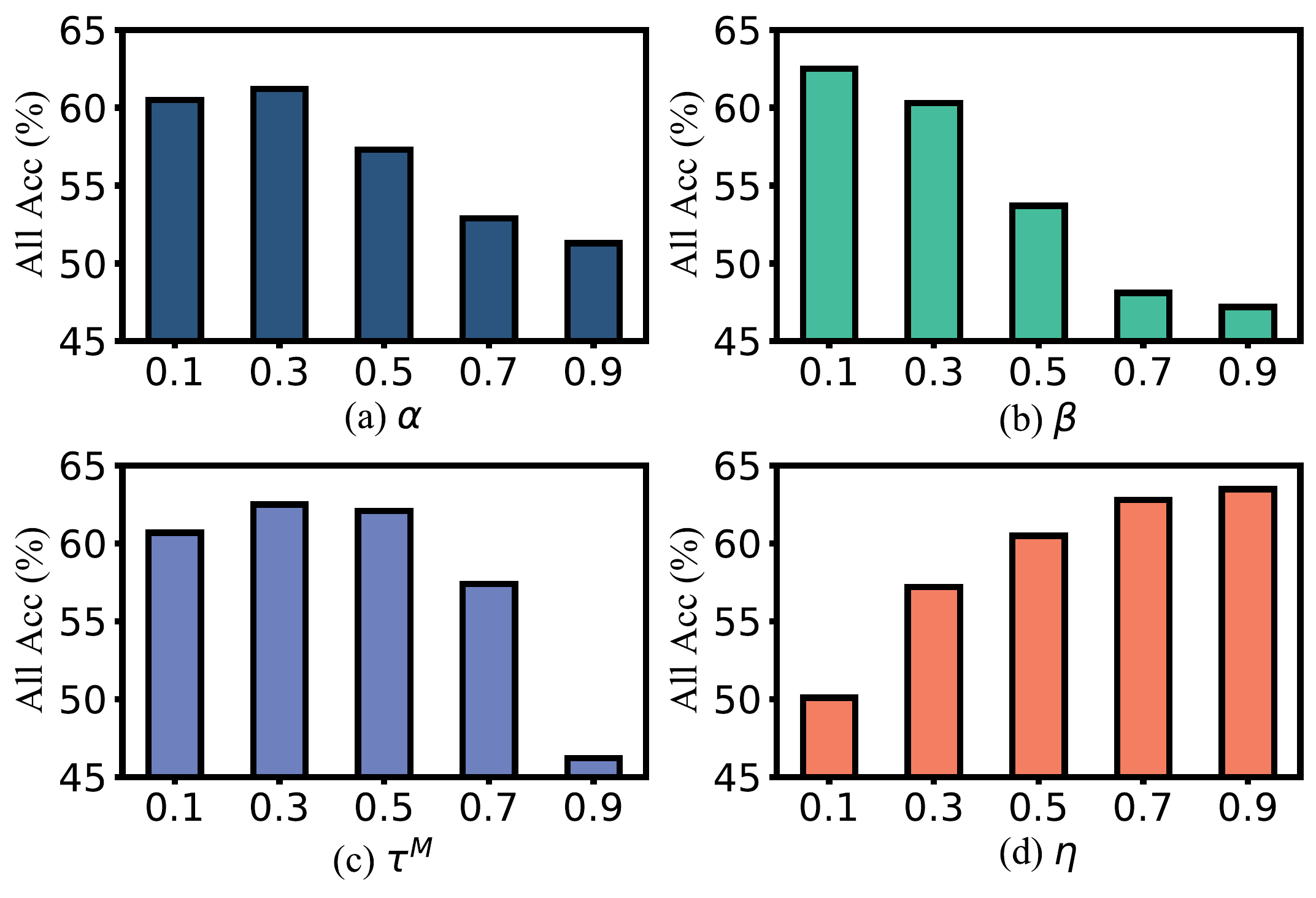}
\caption{Impact of hyper-parameters. The clustering accuracy on ``All'' categories is reported.\label{fig:hyper_parameter}}
  \end{center}
\vspace{-3em}
\end{figure}

\subsection{Hyper-Parameter Analyses}\label{sec:hyper_parameter}
In this section, we discuss the impact of the hyper-parameters in our DCCL, including loss weights ($\alpha$ and $\beta$), the threshold parameter of dispersion loss ($\tau^{M}$), momentum updating factor ($\eta$), and the frequency of DCG ($\tau^{I}$).

\par\noindent
\textbf{Impact of loss weights and threshold parameters.} For the evaluation of loss weights, we use the hold-off validation data to determine their values. Specifically, we first select the optimal $\alpha$ to achieve the best accuracy on the ``All'' score, then we find the optimal $\beta$ based on the selected $\alpha$. The impact of different values is shown in~\cref{fig:hyper_parameter}~(a)-(c). Similarly, we choose the best $\tau^{M}$ with the selected $\alpha$ and $\beta$. Finally, our final model is obtained by using $\alpha$ = 0.3, $\beta$ = 0.1 and $\tau^{M}$ = 0.3.
\par\noindent
\textbf{Impact of Update Rates.} The impact of updating factor of CMB is illustrated in in~\cref{fig:hyper_parameter}~(d). The more smooth running average can obtain better performance. Considering the balance between computational consumption and performance, we thus set $\tau^{I}$ = 5 and $\eta$ = 0.9 in all experiments.

\begin{figure}[t]
  \begin{center}
  \includegraphics[width=0.48\textwidth]{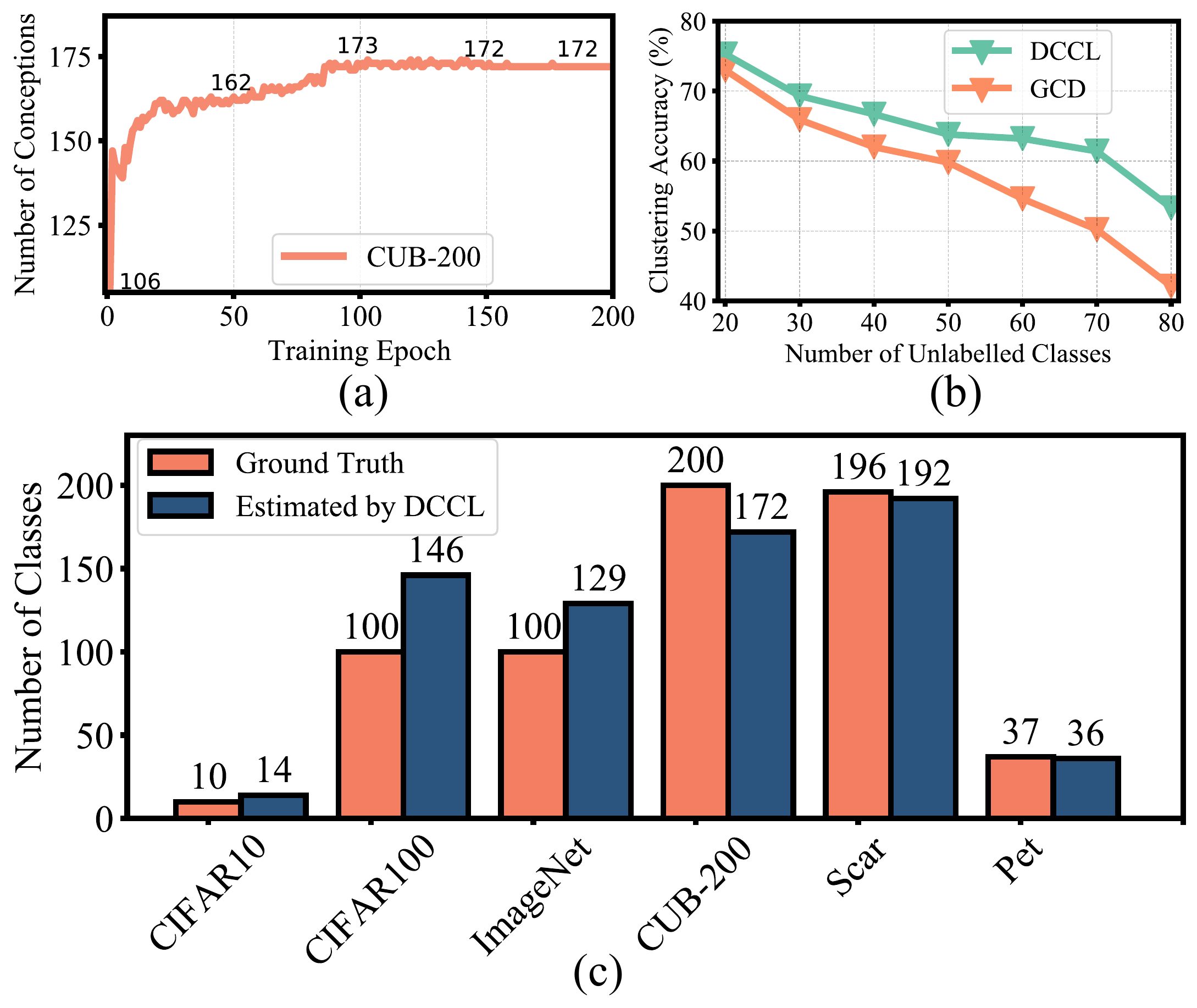}
\caption{(a) visualizes the dynamics of DCG, which implies that our DCG adaptively estimates conceptional representation for different training stages. (b) illustrates the trend of performances on the CIFAR100 dataset, with varying the number of unlabeled classes. (c) is the comparison between the real number of classes in datasets and the estimated number of conceptions.\label{fig:number_cluster}}
  \end{center}
\vspace{-2em}
\end{figure}

\subsection{Further Investigation of DCG}\label{sec:ab_dcg}
To explore our dynamic conception generation (DCG), we conduct two group experiments and the experimental results are shown in~\cref{fig:number_cluster} and~\cref{tab:clustering_method}. From the~\cref{fig:number_cluster}~(a) we find that in the initial training stage, our DCG tends to generate fewer conceptions than at the convergence stage. This is because in the beginning, the model could not understand fine-grained classes well. Thus, DCG generates coarse-grained supervision like super-class, which has a low risk to over-correct. Later, with the growth of feature discriminability, DCG builds elaborate conceptional relationships to further refine the learned representations.
\par
From the~\cref{fig:number_cluster}~(c), we find that the number of conceptions estimated by DCCL is close to the ground truth of the number of classes in the corresponding dataset. However, it is worth noting that the conception representations in this paper are not equivalent to the cluster centers. The DCG aims to adaptively generate proper conceptional representations that are beneficial for contrastive learning, instead of predicting the real number of classes within a dataset. 
Furthermore, we try to directly replace our DGC with semi-supervised $k$-means~\cite{vaze2022generalized} (SSK) with the prior of known $k$. From the results in~\cref{tab:clustering_method}, SSK fails to generate effective latent conceptions, even leading to worse clustering.
\par\noindent
\textbf{Discussion.} Based on the comparison in~\cref{fig:number_cluster}~(b), we notice that when training on generic datasets, our DCG tends to generate more conceptions than the actual number of dataset classes, while generating fewer conceptions on fine-grained datasets. A possible explanation is that due to large inter-class differences in generic datasets, learning more sub-classes conceptions enables models to gain discriminability. Similarly, since fine-grained classes share more common attributes or conceptions, such super-class information can significantly benefit fine-grained GCD.

\begin{table}[!t]
\centering
\caption{Comparison of different clustering methods for DCG.\label{tab:clustering_method}}
\begin{tabular}{l|ccc}
\cmidrule[1pt]{1-4}
\multicolumn{1}{c|}{\multirow{2}{*}{Clustering Method}} &  \multicolumn{3}{c}{CUB-200~\cite{wah2011caltech}} \\ \cmidrule[0.5pt]{2-4}
\multicolumn{1}{c|}{}                                                                           & All    & Old    & New   \\ \cmidrule[0.5pt]{1-4}
$k$-means\cite{macqueen1967classification}                                                                                &       54.1  &  53.4  &  53.3 \\
SSK~\cite{vaze2022generalized}                                                                                     & 55.9  &  55.1  &  54.2     \\
FINCH\cite{sarfraz2019efficient}   &  55.8& 56.1 &55.4 \\
DBSCAN~\cite{ester1996density}                                                                                   &   60.5   &  54.7  & 59.8  \\
InfoMap~\cite{rosvall2008maps}                                                                                   & 61.4    &55.2&62.7\\
DCG (Ours)                                                                          & \bf63.5   &\bf60.8  &\bf64.9\\ \cmidrule[1pt]{1-4}
\end{tabular}
\vspace{-1em}
\end{table}
\subsection{Evaluation with Different Split Protocols}
To explore the effects of DCCL under strict annotation limitation, we propose to test models on varying splits of CIFAR100~\cite{krizhevsky2009learning}. We visualize the accuracy of ``All'' classes in~\cref{fig:number_cluster}~(b), which indicates that our DCCL has stronger robustness when only a few labeled classes are available. Meanwhile, with the growth of the number of unlabeled classes, GCD expresses a severe performance degradation. 
\section{Conclusion}
In this paper, we propose to cope with the generalized category discovery (GCD) from the perspective of mining underlying relationships between known and unknown categories. To implement this idea, we propose a dynamic conceptional contrastive learning framework to alternately explore latent conceptional relationships and perform conceptional contrastive learning. This mechanism enables models to learn more discriminable representations. Furthermore, to mitigate the inconsistency of updating conception representations during the training process, we propose a conception-level momentum update to facilitate the model toward better optimization. Extensive experimental results show that our DCCL achieves a new state-of-the-art performance on GCD tasks.

\noindent\textbf{Acknowledgement} This work has been supported by the EU H2020 project AI4Media (No.
951911) and by the PRIN project CREATIVE (Prot. 2020ZSL9F9).

{\small

\bibliographystyle{ieee_fullname}
\bibliography{egbib}
}

\end{document}